\documentclass[10pt,twocolumn,letterpaper]{article}

\usepackage[
]{cvpr}      

\usepackage{graphicx}
\usepackage{amsmath}
\usepackage{amssymb}
\usepackage{booktabs}
\usepackage{amsmath} 
\usepackage{appendix}
\usepackage{enumitem}
\usepackage[pagebackref,breaklinks,colorlinks]{hyperref}

\usepackage[capitalize]{cleveref}
\crefname{section}{Sec.}{Secs.}
\Crefname{section}{Section}{Sections}
\Crefname{table}{Table}{Tables}
\crefname{table}{Tab.}{Tabs.}
\def\cvprPaperID{*****} 
\def\confName{CVPR}
\def\confYear{2022}

\begin{document}

\title{Comparative Analysis of Deep Convolutional Neural Networks \\for Detecting Medical Image Deepfakes}

\author{Abdel Rahman Alsabbagh$^{1,2}$ \quad Omar Al-Kadi$^{1}$\\
$^{1}$University of Jordan, Jordan\\
$^{2}$King Abdullah University of Science and Technology, Saudi Arabia\\
\tt\small abdelrahman.sabbagh@kaust.edu.sa \quad \tt\small o.alkadi@ju.edu.jo
}
\maketitle

\begin{abstract}
   Generative Adversarial Networks (GANs) have exhibited noteworthy advancements across various applications, including medical imaging. While numerous state-of-the-art Deep Convolutional Neural Network (DCNN) architectures are renowned for their proficient feature extraction, this paper investigates their efficacy in the context of medical image deepfake detection. The primary objective is to effectively distinguish real from tampered or manipulated medical images by employing a comprehensive evaluation of 13 state-of-the-art DCNNs. Performance is assessed across diverse evaluation metrics, encompassing considerations of time efficiency and computational resource requirements. Our findings reveal that ResNet50V2 excels in precision and specificity, whereas DenseNet169 is distinguished by its accuracy, recall, and F1-score. We investigate the specific scenarios in which one model would be more favorable than another. Additionally, MobileNetV3Large offers competitive performance, emerging as the swiftest among the considered DCNN models while maintaining a relatively small parameter count. We also assess the latent space separability quality across the examined DCNNs, showing superiority in both the DenseNet and EfficientNet model families and entailing a higher understanding of medical image deepfakes. The experimental analysis in this research contributes valuable insights to the field of deepfake image detection in the medical imaging domain\footnote{\scriptsize{\MakeUppercase{Technical Report, University of Jordan, Artificial Intelligence Department, TR-01,23.}}}.
\end{abstract}

\section{Introduction}
\label{sec1}

In the realm of medical imaging, the advent of generative modeling marks a transformative era. Traditional data augmentation techniques, effective in many domains, encounter limitations when applied to medical images like Computed Tomography (CT) scans or Magnetic Resonance (MR) scans. Geometric transformations, such as random flipping, cropping, rotation, or translation, prove inadequate, failing to enhance neural network generalization capabilities beyond the initial training population and often resulting in the generation of highly correlated samples \cite{Garcea2023}.

Recognizing these challenges, recent strides have been made in leveraging Generative Adversarial Networks (GANs) as a solution \cite{Osuala2021, 2021paired}. GANs contribute by generating authentic-looking medical images, augmenting datasets, and positively impacting model accuracy. This not only simplifies data synthesis within the medical imaging domain but also offers a cost-effective alternative. However, the application of GANs introduces challenges, such as incorporating intentional manipulation, forgery, and tampering in the medical images, potentially giving rise to future apprehensions among clinicians considering the integration of AI in the field of medicine. Authenticating the generated images is crucial, given the potential consequences of misinterpretation regardless of the intent behind the application. To this end, this paper revolves around the utilization of state-of-the-art Deep Convolutional Neural Network (DCNN) architectures to discern between authentic and synthetic CT scan images, generated by the CT-GAN \cite{Mirsky2019}. The hypothesis driving our investigation is rooted in the critical need for a reliable approach to authenticate medical images, especially in scenarios where the visual realism of GAN-generated images poses challenges to accurate diagnosis.
Our contributions are:
\begin{enumerate}[label=(\alph*)]
    \item Conducting a set of extensive experiments on the most prominent DCNNs known to be used by the machine learning community for medical image deepfake detection tasks;
    \item Analyzing time complexity and model efficiency for a finer understanding of DCNNs in the medical domain;
    \item Exploring the embeddings of various DCNNs after training on the medical image deepfake detection task, and examining the latent space separability quality. 
\end{enumerate}

Through this paper, we present a comprehensive overview of the study and related work in section \ref{overviewsec}. Subsequently, we detail the dataset preprocessing, methodology, and results analysis in subsequent sections (\ref{datapresec}, \ref{methodsec}, and \ref{resultssec}), providing valuable insights into the potential of DCNNs in addressing authenticity challenges in medical imaging. Finally, section \ref{concsec} further discuss and summarize our findings and conclusion.

\section{Overview}\label{overviewsec}
The field of deepfake detection has experienced a surge in research activity, driven by milestones like the introduction of StyleGAN \cite{Karras2018}. This Generative Adversarial Network (GAN) marked a transformative shift by excelling in the creation of highly realistic images, surpassing its predecessors. Researchers have responded to the challenge of detecting forgery in such images through a diverse array of approaches. These range from machine learning-based methods, such as tree-based methods \cite{Rana2021} claiming enhanced interpretability, to the utilization of vanilla Convolutional Neural Networks (CNNs) \cite{Badale2018}.

The adoption of well-established or novel Deep Convolutional Neural Network (DCNN) architectures has been a prevalent theme in recent research efforts. For instance, \cite{Khatri2023} utilized VGG16 \cite{simonyan2014very}, MobileNetV2 \cite{sandler2018mobilenetv2}, Xception \cite{chollet2017xception}, and InceptionV3 \cite{szegedy2016rethinking} to detect deepfakes on generic data, \cite{Rana2020} employed a combination of ensemble learning techniques with CNNs, and \cite{Chen2021} developed a lightweight deepfake detector using the successive subspace learning principle. Some researchers have explored hand-crafted features, considering biological signals within images \cite{Ciftci2019}.

Beyond image manipulation, the creation of deepfakes has extended to videos and audio, necessitating the application of Recurrent Neural Networks (RNNs) to model forgery in sequential data. \cite{Chen2020} addressed audio deepfakes using a large margin cosine loss function and frequency masking, \cite{Guera2018} used recurrent neural networks for video deepfake detection, and \cite{Zhou2021} and \cite{Mittal2020} developed methods to detect joint audio-visual deepfakes. While Transformers have revolutionized natural language processing \cite{Vaswani2017} and extended to images \cite{Dosovitskiy2020}, attention-based blocks have been introduced in various deepfake detection techniques. \cite{Wang2022}, for instance, utilized the Transformer to detect multimodal deepfakes on multiple scales, and \cite{Ganguly2022} assembled a hybrid Transformer-CNN architecture to capture both local and global contexts within an image.

As the field of deepfake detection matures, it finds novel applications in the medical imaging domain, addressing unique challenges. So far, \cite{Budhiraja2022} employed convolutional reservoir networks, and \cite{Solaiyappan2022} conducted a comparative study on a limited set of machine learning and deep learning architectures. Our work is inspired by this evolving landscape, with a specific focus on the capabilities of DCNNs. Unlike previous works, our concentration on DCNNs aims to discern differentiating features in medical images, providing a deeper understanding of their capabilities in capturing complex features within the medical imaging domain.

In the following sections, we delve into the methodologies employed, dissect encountered obstacles and vulnerabilities, and analyze the solutions proposed to overcome challenges in our exploration of DCNNs for detecting deepfakes in medical imaging. This comprehensive overview critically examines state-of-the-art approaches, laying the foundation for subsequent sections in our work.

\begin{figure*}[t]%
\centering
\includegraphics[width=\textwidth]{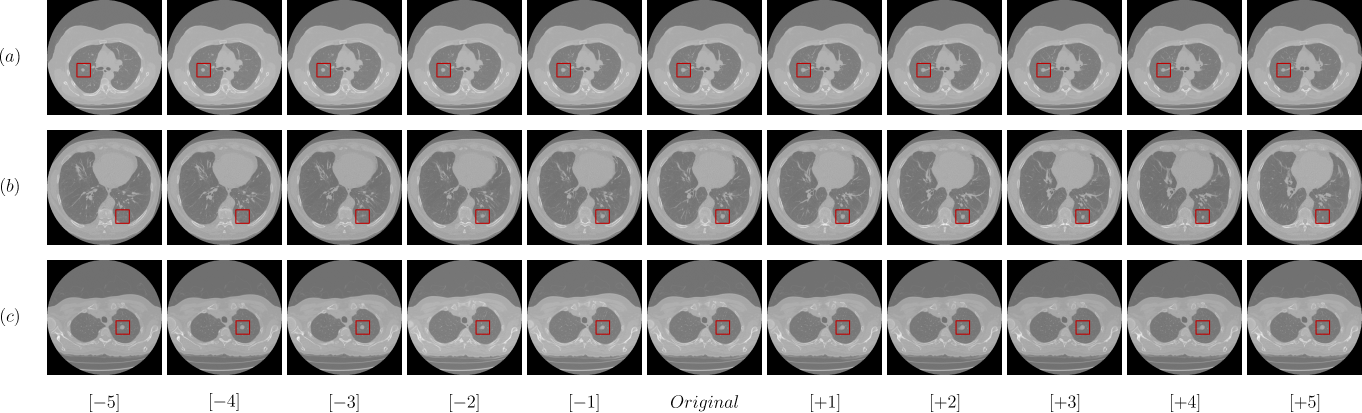}
\caption{Sequential scan slices of three patients denoted as (a), (b), and (c) on each row, where the column \textit{Original} denotes the label provided, and columns $[\pm \textit{i}]$ denote the adjacent slices within a range of \textit{i} of the \textit{Original} slice. (a) shows that the tumor starts and ends exactly within the range $[-5, +5]$, (b) shows that the tumor starts and ends within the range $[-3, +4]$, and finally  (c) shows that the tumor starts and ends out of the range $[-5, +5]$.}\label{comparison}
\end{figure*}

\section{Data Preprocessing}\label{datapresec}

In this section, we will discuss both the datasets used and how we preprocessed the data before employing it in our models.

\subsection{Datasets}\label{subsec3.1}
In constructing a model capable of discerning between \textit{real} and \textit{fake} medical images, the utilization of diverse and representative datasets is imperative. For the \textit{real} medical image dataset, we employed the Lung Image Database Consortium image collection (LIDC-IDRI), a comprehensive repository of lung CT scans encompassing data from a thousand distinct patients \cite{Armato2011}. This dataset serves as a benchmark for authentic medical images, providing a diverse and well-established foundation for our model's understanding of genuine medical scans.

Conversely, to introduce the dimension of \textit{fake} medical images, we leveraged tampered data generated by the CT-GAN. Notably, the CT-GAN utilized the LIDC-IDRI dataset during its training phase, making it an intriguing source for simulated or manipulated medical images. This dual dataset approach, incorporating both authentic and manipulated medical images, enables our model to learn discriminative features essential for distinguishing between \textit{real} and \textit{fake} instances. Through this balanced and diverse dataset selection, we aim to enhance the robustness and generalization capabilities of our deep learning model in the domain of medical image deepfake detection.

\subsection{Preprocessing}\label{subsec3.2}
Both datasets contain files in Digital Imaging and Communications in Medicine (DICOM) format, and it is a widely used standard for storing and transmitting medical images. The format includes not only the image itself but also the metadata of the scan such as the diagnosis of the scan and the patient's information. A medical scan folder contains a stack of 2D image slices in DICOM format that can be reconstructed into a 3D volume.
The CT-GAN dataset has a total of a hundred scans, each scan is a series of $512 \times 512$ images, and the series ranges from $100 - 300$ slices long. In the metadata file provided by \cite{Mirsky2019}, there were only 113 slices labeled as \textit{false malignant} and \textit{false benign}, i.e. \textit{tampered / fake}. It appears to suffer from severe imbalance, but the CT-GAN model injects and removes nodules in a 3D volume and not only one slice of the whole scan. So in principle, the neighboring slices should also extend the nodule of a labeled slice, as this is how the 3D pix2pix network works \cite{Isola2017}.

To examine our preprocessing approach, we visualized the labeled slices and the adjacent ones within different ranges. Figure~\ref{comparison} shows three separate \textit{False Malignant} scans with their tumor region with a range of $[-5, +5]$ of the original label. We can see that indeed the tumor does not show up only on the labeled slice, but also on the neighbor slices, which means in return that there are more than 113 tampered images.
At this stage, there were two choices to get a hold of as much tampered image data as possible: a) visualize all labeled slices and their adjacent slices and select manually all of the slices that include a tumor at the region of interest, or b) automatically select the labeled slices and their adjacent slices within a range of  $[-5, +5]$ of the labeled slice. The initial option is deemed impractical; therefore, the latter is chosen. However, it is not a strict rule that all slices within the range of $[-5, +5]$ of the labeled slice are consistently tampered, as illustrated in Figure~\ref{comparison}. We acknowledge the potential for errors associated with this choice. Nevertheless, in a significant majority of cases, nearly all adjacent slices within the range $[-5, +5]$ do exhibit tampering. After this preprocessing step, we ended up having $1,243$ images for the \textit{fake} class.

The LIDC-IDRI dataset contains scans of a thousand patients with precisely $244,385$ scan slices for all patients. To maintain a balanced overall dataset with a $50{:}50$ \textit{real:fake} ratio, we down-sampled the LIDC-IDRI dataset to have $1,243$ images, which was achieved by choosing slices at random, making the overall dataset that will be used in training a total of $2,486$ images. 

\section{Methodology}\label{methodsec}

To solve the medical deepfake detection task, we experimented with a diverse set of DCNN architectures, these include ConvNeXtTiny \cite{liu2022convnet}, DenseNet121, DenseNet169, DenseNet201 \cite{huang2017densely}, EfficientNetB4 \cite{tan2019efficientnet}, EfficientNetV2S \cite{tan2021efficientnetv2}, InceptionV3 \cite{szegedy2016rethinking}, MobileNetV3Large \cite{howard2019searching}, RegNetX040, RegNetY040 \cite{radosavovic2020designing}, ResNet50V2 \cite{he2016identity}, VGG19 \cite{simonyan2014very}, and Xception \cite{chollet2017xception}. The rationale behind selecting these models particularly is detailed in Appendix \ref{selection}.

\begin{figure}
\centering
\includegraphics[width=\columnwidth]{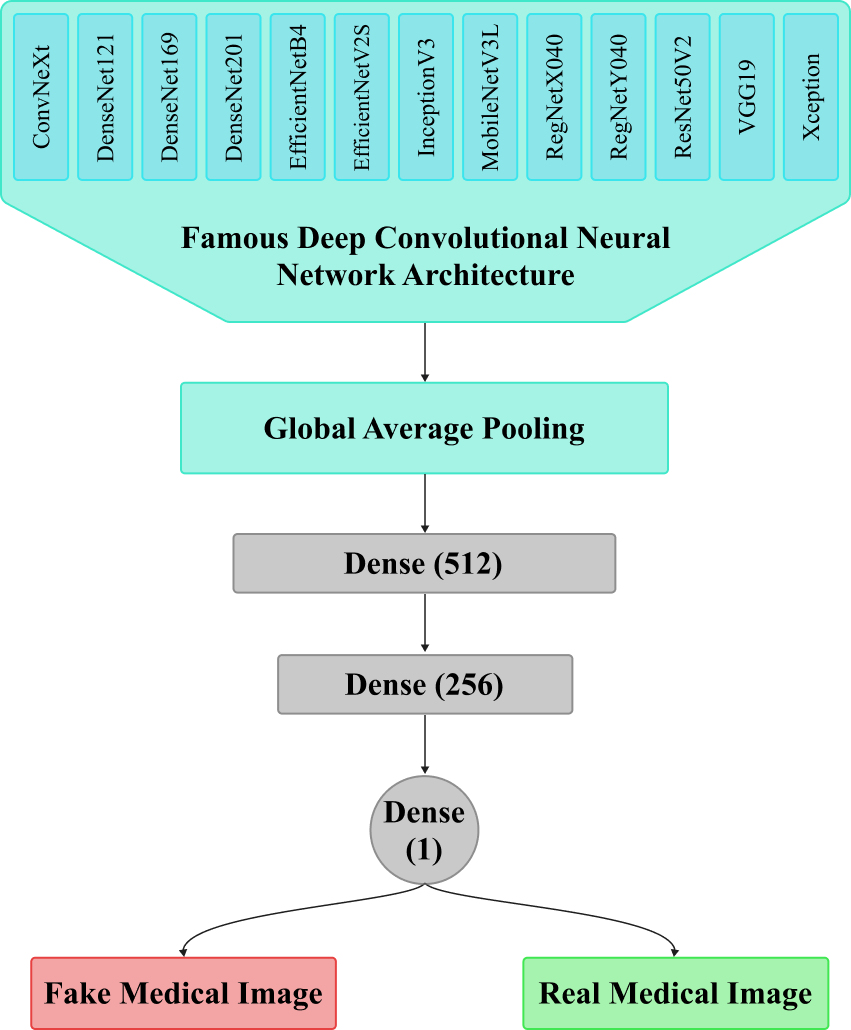}
\caption{Full architecture used for our experiments, which consists of a Deep Convolutional Neural Network (DCNN), a global average pooling layer, a Dense (fully-connected) layer of size 512 followed by a size of 256, and finally a Sigmoid activation function indicating a probability of how much an image is \textit{real}.}\label{dcnn}
\end{figure}

We modified the DCNNs by appending a global average pooling layer at the end, followed by a couple of fully connected layers as additional feature extractors tailored to the binary classification problem at hand and for the later latent space visualization. The full architecture is depicted in Figure \ref{dcnn}. 

\begin{table*}[t]
\caption{Results of different Deep Convolutional Neural Networks (1 run)}\label{resultstable}
\begin{tabular*}{\textwidth}{@{\extracolsep\fill}lcccccc}
\toprule%
Model & Accuracy & Precision & Recall/Sensitivity & Specificity & F1-Score & AUC \\
\midrule
ConvNeXtTiny & 0.9544 & 0.9444 & 0.9605 & 0.9490 & 0.9524 & 0.9968 \\
DenseNet121 & 0.9678 & 0.9412 & \textbf{0.9944} & 0.9439 & 0.9671 & \textbf{0.9988} \\
DenseNet169 & \textbf{0.9786} & 0.9669 & 0.9887 & 0.9694 & \textbf{0.9777} & 0.9985 \\
DenseNet201 & 0.9571 & 0.9448 & 0.9661 & 0.9490 & 0.9553 & 0.9971 \\
EfficientNetB4 & 0.9678 & 0.9661 & 0.9661 & 0.9694 & 0.9661 & 0.9980 \\
EfficientNetV2S & 0.9678 & 0.9609 & 0.9718 & 0.9643 & 0.9663 & 0.9980 \\
InceptionV3 & 0.9517 & 0.9344 & 0.9661 & 0.9388 & 0.9500 & 0.9968 \\
MobileNetV3Large & 0.9598 & 0.9451 & 0.9718 & 0.9490 & 0.9583 & 0.9975 \\
RegNetX040 & 0.9491 & 0.9593 & 0.9322 & 0.9643 & 0.9456 & 0.9949 \\
RegNetY040 & 0.9544 & 0.9348 & 0.9718 & 0.9388 & 0.9529 & 0.9945 \\
ResNet50V2 & 0.9732 & \textbf{0.9883} & 0.9548 & \textbf{0.9898} & 0.9713 & 0.9982 \\
VGG19 & 0.9517 & 0.9492 & 0.9492 & 0.9541 & 0.9492 & 0.9964 \\
Xception & 0.9651 & 0.9659 & 0.9605 & 0.9694 & 0.9632 & 0.9977 \\
\bottomrule
\end{tabular*}
\end{table*}

\begin{table*}[t]
\caption{Results of top-performing Deep Convolutional Neural Networks (3 runs)}\label{winresultstable}
\small
\begin{tabular*}{\textwidth}{@{\extracolsep\fill}lcccccc}
\toprule%
Model & Accuracy & Precision & Recall/Sensitivity & Specificity & F1-Score & AUC \\
\midrule
DenseNet121 & 0.9705 $\pm$0.0047 & 0.9547 $\pm$0.0153 & 0.9849 $\pm$0.0214 & 0.9592 $\pm$0.0216 & 0.9694 $\pm$0.0050 & \textbf{0.9984 $\pm$0.0004} \\

DenseNet169 & \textbf{0.9732 $\pm$0.0047} & 0.9500 $\pm$0.0147 & \textbf{0.9962 $\pm$0.0007} & 0.9567 $\pm$0.0180 & \textbf{0.9738 $\pm$0.0055} & 0.9982 $\pm$0.0003 \\

EfficientNetB4 & 0.9642 $\pm$0.0040 & 0.9507 $\pm$0.0137 & 0.9755 $\pm$0.0086 & 0.9541 $\pm$0.0135 & 0.9623 $\pm$0.0040 & 0.9940 $\pm$0.0062 \\

EfficientNetV2S & 0.9615 $\pm$0.0086 & 0.9502 $\pm$0.0109 & 0.9699 $\pm$0.0086 & 0.9541 $\pm$0.0102 & 0.9599 $\pm$0.0090 & 0.9976 $\pm$0.0005 \\

ResNet50V2 & 0.9723 $\pm$0.0016 & \textbf{0.9746 $\pm$0.0288} & 0.9680 $\pm$0.0279 & \textbf{0.9762 $\pm$0.0281} & 0.9710 $\pm$0.0001 & 0.9981 $\pm$0.0001 \\

Xception & 0.9418 $\pm$0.0295 & 0.9246 $\pm$0.0538 & 0.9579 $\pm$0.0029 & 0.9274 $\pm$0.0560 & 0.9404 $\pm$0.0284 & 0.9936 $\pm$0.0066 \\
\bottomrule
\end{tabular*}
\end{table*}

All experiments were conducted using Google Colab's NVIDIA T4 GPU. We loaded the DCNNs directly from the \texttt{TensorFlow} \cite{tensorflow2015-whitepaper} and initiated them with pre-trained weights from the ImageNet dataset \cite{Deng2009imagenet}, keeping the base models frozen while allowing the remaining architecture to be trainable. The input image size was fixed at $512 \times 512$ pixels, and we uniformly trained all models for 80 epochs, saving models with the lowest validation loss. A batch size of 8 was used, and we employed Binary Cross-Entropy (BCE) as the loss function as follows:
\begin{equation}
\text{BCE} = -\frac{1}{N}\sum_{i=1}^{N} \left[ y_i \log(\hat{y}_i) + (1 - y_i) \log(1 - \hat{y}_i) \right]
\end{equation}
where $N$ is the number of samples in the dataset, $y_i$ is the true binary label for the $i$-th sample, and $\hat{y}_i$ is the predicted probability that the $i$-th sample belongs to the \textit{real} class.\\
We used Adam \cite{kingma2014adam} as an optimizer with a learning rate of $10^{-4}$, $\beta_1$ set to 0.9, and $\beta_2$ set to 0.999.
Following the initial training, we unfroze the base models and fine-tuned them for an additional 20 epochs with a learning rate of $10^{-5}$. Once again, we saved the models with the least validation loss, following the same protocol as in the initial training phase. To evaluate the performance of our model, we used several evaluation metrics, including accuracy, precision, recall, specificity, F1-score, and area under the curve (AUC). Extensive details about the metrics are in Appendix~\ref{eval}.

\section{Results \& Discussion}\label{resultssec}
The experiments involved the execution of a binary classification task designed to discriminate between medical images categorized as \textit{real} and \textit{fake} across a collection of 13 distinct models. In the initial phase, we conducted a single round of evaluation, wherein we computed the evaluation metrics (defined in Appendix \ref{eval}), after which we calculated the harmonic mean of prominent performance metrics, including accuracy, F1-Score, and specificity, for each of the models. Subsequently, we employed this assessment to identify the top-performing six models, subjecting them to two additional evaluation iterations. This way we not only expedited the process of model selection but also facilitated a more comprehensive investigation of their stability characteristics.

\paragraph{ResNet50V2 Exhibits Distinctive Performance} With consistent robustness in comparison to other models.
This distinction is evident in the evaluation metrics detailed in Appendix \ref{eval}, summarized in both Table \ref{resultstable} and Table \ref{winresultstable}. ResNet50V2 particularly excels, achieving the highest precision and recall, both around 99\%.

\begin{figure}
\centering
\includegraphics[width=\columnwidth]{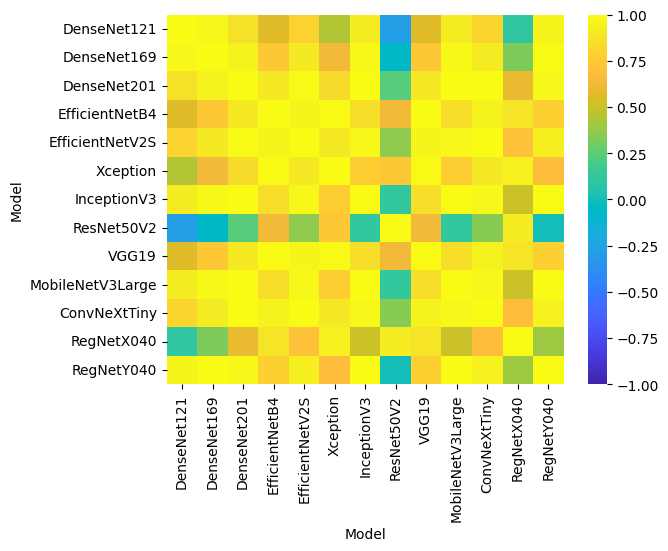}
\caption{Pearson's correlation matrix showing interrelationships among Deep Convolutional Neural Networks (DCNNs) results.}\label{resultsfig}
\end{figure}

\begin{figure}[t]
\centering
\includegraphics[width=\columnwidth]{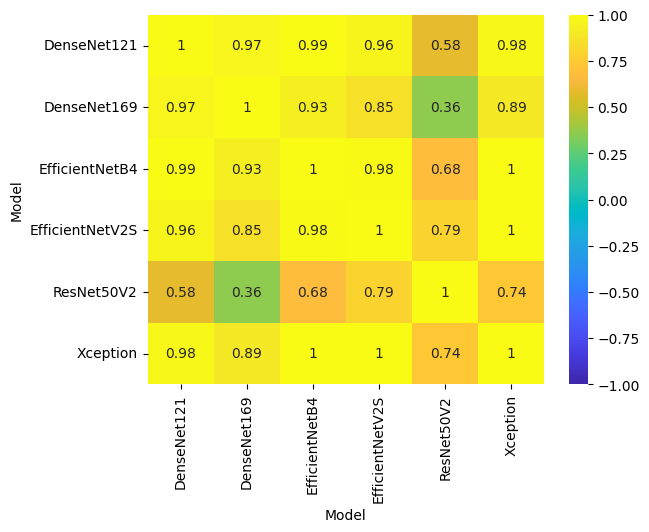}
\caption{Pearson's correlation matrix for the top-performing Deep Convolutional Neural Networks results.}\label{bestresultsfig}
\end{figure}

Figures \ref{resultsfig} and \ref{bestresultsfig} visually highlight ResNet50V2's unique characteristics, as its outcomes exhibit a notable lack of substantial correlation with those of other models. This is measured using Pearson's correlation coefficient, defined as:
\begin{equation}
\rho =  \frac{\sum(x_i - \Bar{x})(y_i - \Bar{y})}{\sqrt{\sum(x_i - \Bar{x})^2(y_i - \Bar{y})^2}}\label{r}
\end{equation}
where $x$ and $y$ represent the results of each model in vector form, constructed as $[\text{Accuracy}_m, \text{Precision}_m, \text{Recall}_m, \text{Specificity}_m, \text{F1-Score}_m,$ $\text{AUC}_m]^T$.

\begin{figure}
\includegraphics[width=\columnwidth]{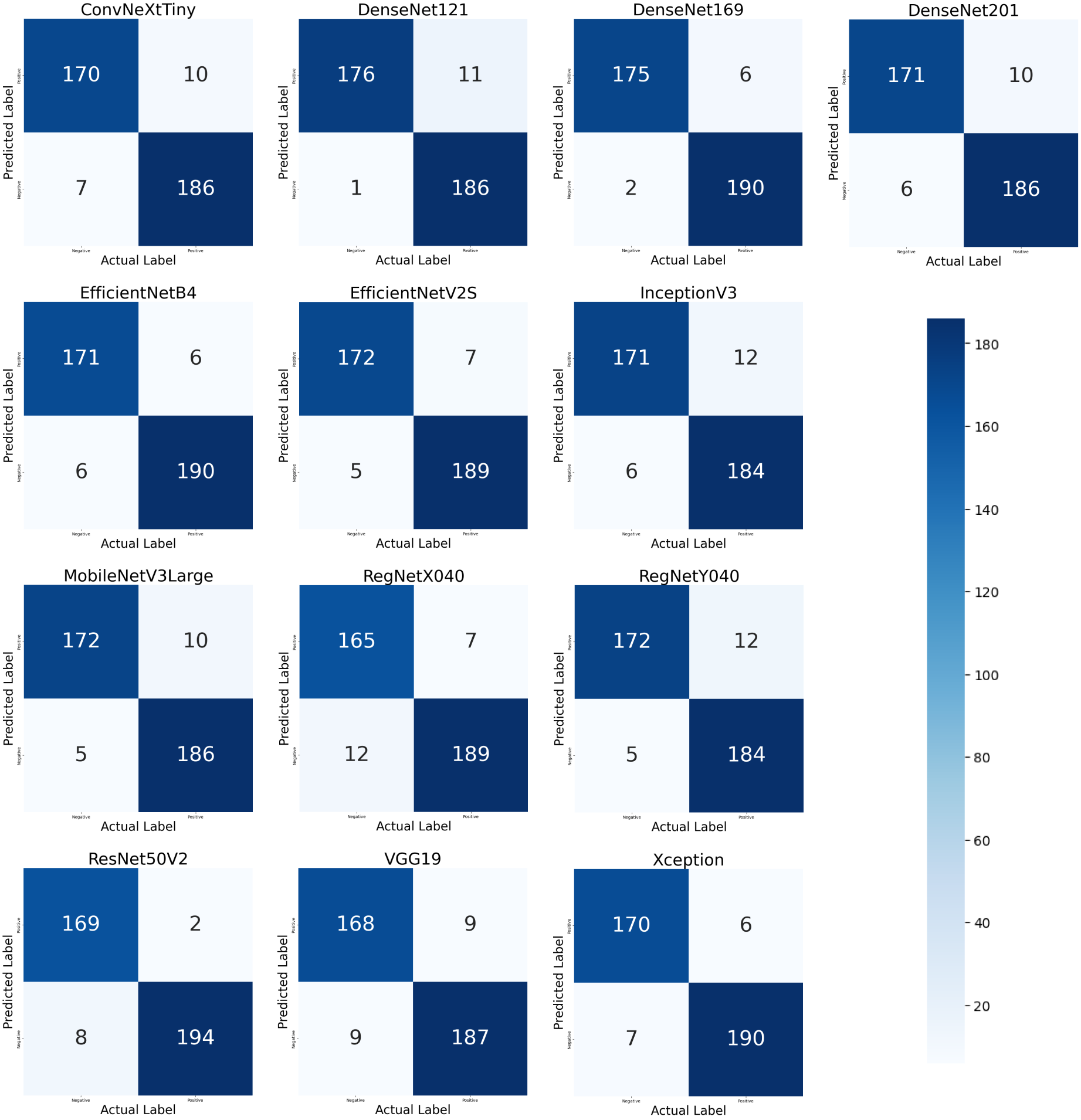}
\caption{Confusion matrices of Deep Convolutional Neural Networks.}\label{ConfusionMatrix}
\end{figure}

This distinct performance of ResNet50V2 is attributed to the intrinsic characteristics of the residual blocks within the ResNet family. These blocks effectively preserve features from the input image and consistently integrate them across subsequent layers, contributing to the model's robustness. It is noteworthy to observe that all DCNNs demonstrated commendable proficiency in capturing differentiating features, as illustrated in Figure \ref{ConfusionMatrix}. 

\begin{table}
\caption{Number of parameters and mean inference step time of different Deep Convolutional Neural Networks (10 runs) }\label{dcnntable}
\small
\begin{tabular*}{\columnwidth}{@{\extracolsep\fill}lcc}
\toprule%
Model & Parameters (M) & Inference step (ms)\\
\midrule
ConvNeXtTiny & 28.3 & 582.1 $\pm 18.6$ \\
DenseNet121 & 7.7 & 334.9 $\pm 72.4$ \\
DenseNet169 & 13.6 & 396.9 $\pm 67.2$ \\
DenseNet201 & 19.4 & 409.2 $\pm 51.8$ \\
EfficientNetB4 & 18.7 & 442.9 $\pm 52.7$ \\
EfficientNetV2S & 21.1 & 382.6 $\pm 65.7$ \\
InceptionV3 & 23.0 & 380.7 $\pm 86.6$ \\
MobileNetV3Large & \textbf{3.6} & \textbf{325.4 $\pm$44.7} \\
RegNetX040 & 21.7 & 457.2 $\pm 56.1$ \\
RegNetY040 & 20.3 & 446.3 $\pm 86.3$ \\
ResNet50V2 & 24.7 & 368.3 $\pm 87.2$ \\
VGG19 & 20.4 & 490.2 $\pm 45.1$ \\
Xception & 22.0 & 395.3 $\pm 76.9$ \\
\bottomrule
\end{tabular*}
\end{table}

\begin{figure}[t]
\centering
\includegraphics[width=\columnwidth]{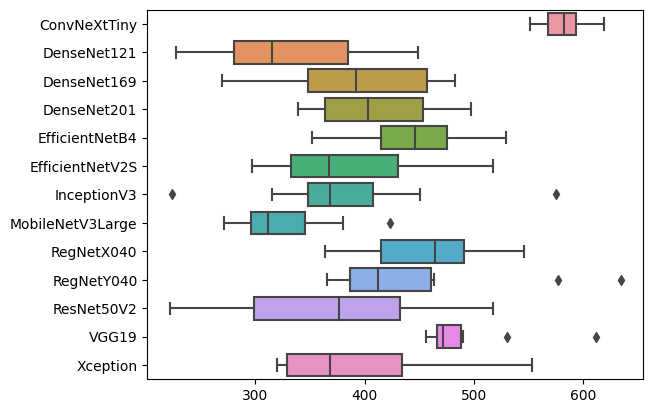}
\caption{Time recorded per inference step after ten inferences on various Deep Convolutional Neural Networks}\label{timeboxplot}
\end{figure}
\begin{figure*}
    \centering
    \includegraphics[width=\textwidth]{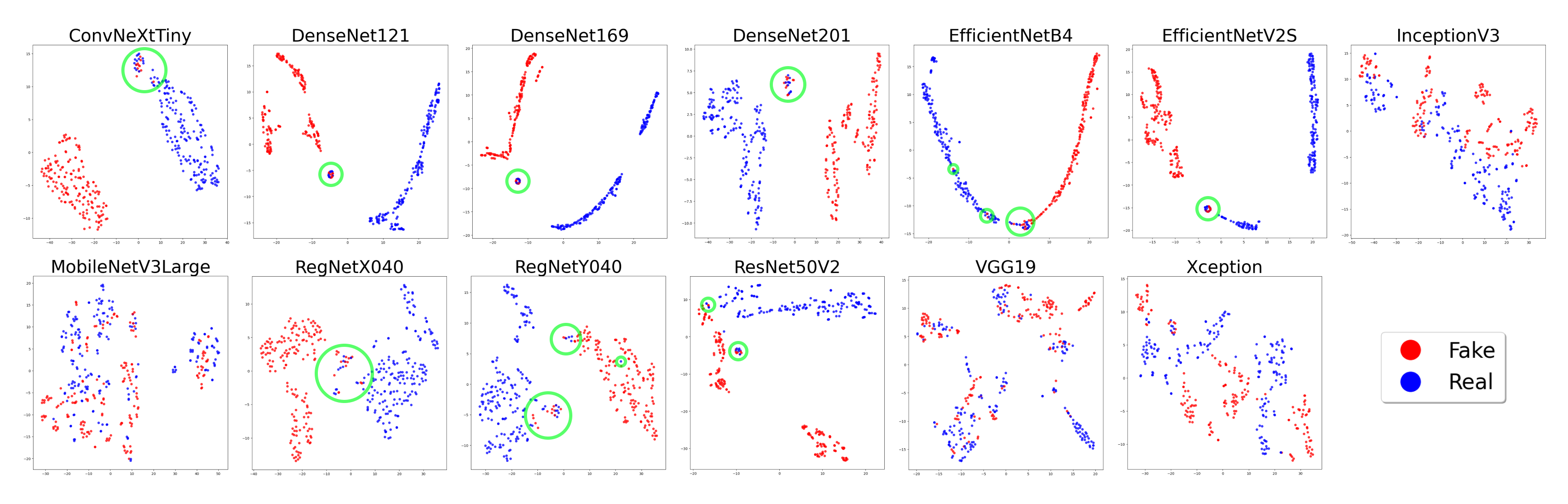}
    \caption{Latent space separability quality between \textit{real} and \textit{fake} examples, showing penultimate layer embedding of each Deep Convolutional Neural Network (DCNN). (minor areas of collision circled in green)}
    \label{TSNE}
\end{figure*}

\paragraph{DenseNet169 Consistently Excels Across Metrics} Achieving the highest levels of accuracy, recall, and F1-score, as observed over three iterative runs. This consistent excellence underscores its stability and efficacy in addressing the non-uniform distribution of tumor types and nodules within CT scan images. Comparatively, DenseNet121 demonstrates a performance comparable to DenseNet169, achieving similar results with nearly half the number of parameters and quicker inference times. These findings are summarized in Table \ref{dcnntable} and visually presented in Figure \ref{timeboxplot}. Notably, this suggests that DenseNet121 offers a favorable trade-off between model complexity and performance. However, DenseNet201 exhibits a degradation in performance compared to its counterparts in the DenseNet family. This departure from the expected trend suggests that the conclusions drawn in \cite{Rajpurkar2017} may not readily extend to our specific case. We attribute this discrepancy to potential overfitting, likely exacerbated by the increased complexity of DenseNet201's architecture.

\paragraph{Comparable Efficiency of MobileNetV3Large} Exhibiting remarkable efficiency, and boasting minimal parameters and inference time while delivering similar results to other DCNNs. With approximately 6.81 times and 3.78 times fewer parameters than ResNet50V2 and DenseNet169, respectively, MobileNetV3Large maintains a maximum difference of only around 4\% in comparison to the highest values across all evaluation metrics. This establishes MobileNetV3Large as a promising option, particularly when facing hardware limitations. Conversely, VGG19 demonstrates fair stability in its inference time, aligning with the consistency observed in ConvNeXtTiny and MobileNetV3Large, as illustrated in Figure \ref{timeboxplot}. On the contrary, ResNet50V2 exhibits notable inconsistency in inference time. 

\paragraph{Distinguished Latent Space Separability of DenseNet169} 
Showing high quality separability between \textit{real} and \textit{fake} data points as seen in Figure \ref{TSNE} with visualization of the second-to-last layer embeddings. Similarly, this is seen in other members of the DenseNet family, particularly in DenseNet121. Interestingly, EfficientNetB4, EfficientNetV2S, and ResNet50V2 have a likewise outstanding ability to separate the different classes. The DCNNs mentioned above all share a common property in their latent space which is the \textit{dense} concentration across the embeddings of the same class. This trait is not seen in the other DCNNs, which tells us that these models are less appropriate for the task of medical image deepfake detection. 
Moreover, there is a phenomenon seen across all DCNNs, which is what we like to call the \textit{areas of collision}, in which examples from opposing classes overlap at different regions in the latent space.

Our exploration of various DCNN architectures has unveiled nuanced differences in their performance, prompting the question: In what context is a particular DCNN most suitable?

In medical image deepfake detection, the relationship between model architecture, complexity, and performance is intricate. For maintaining robust performance resilient to noise and out-of-distribution changes, ResNet50V2 emerges as a top choice. Its near-perfect precision and specificity are attributed to its \textit{residual} connections, which mitigate the vanishing gradients problem, facilitate deep network training, encourage feature reuse, and enhance model expressiveness. This outstanding property is seen in Figures \ref{resultsfig} and \ref{bestresultsfig}, where the Pearson correlation's activations are merely shown in ResNet50V2. However, considering ResNet50V2's shortcomings in other evaluation measures, the DenseNet family, especially DenseNet169, proves valuable. Its superior accuracy, recall, and F1-score indicate a comprehensive and sensitive approach to identifying relevant cases and addressing imbalanced class distributions effectively. The unique \textit{dense} connectivity in DenseNet, promoting feature reuse and efficient gradient flow, is a key factor behind its improved performance.

Despite the distinctive strengths of ResNet50V2 and DenseNet169, they face challenges in meeting time and hardware constraints. MobileNetV3Large addresses these limitations by offering comparable performance while executing the fastest. Designed for optimal performance on mobile phone CPUs, MobileNetV3Large utilizes a blend of hardware-aware network architecture search (NAS) and the NetAdapt \cite{yang2018netadapt} algorithm, contributing to its efficient execution. One thing is that important to mention is that the time complexity is not merely dependent on the architecture and number of parameters, implementation efficiency also plays a pivotal role, which can be the reason why a DCNN like ConvNeXtTiny has a substantially higher inference time while having a little increase of parameters over ResNet50V2.

These trade-offs between efficiency, stability, and performance offer valuable insights for selecting models tailored to specific medical image deepfake detection tasks under varying restrictions. One intriguing prospect is the development of hybrid architectures that combine DCNNs capturing different semantics, aiming to maintain collective efficacy when prioritizing reliable performance, albeit with a less complex model and a subtle loss of performance. 

Moreover, effectively learning class-differentiating features is a complicated task that all DCNNs failed to fully succeed in as previously mentioned.  We can infer from the \textit{areas of collision} (see Figure \ref{TSNE}) that DCNNs can either have intrinsic difficulty in learning dissimilar features that map each data point to its label or that the deepfakes are sufficiently \textit{real}-looking, even to the machine. Another reason can be due to the data mislabeling probability as mentioned in Section \ref{datapresec} or because of the embedding size that can be inadequate to fully capture all the features within a given image.

\section{Conclusion}\label{concsec}

In this paper, we addressed the pivotal challenge of authenticating medical images amidst the complexities introduced by deepfakes. Employing a diverse range of DCNN architectures, we utilized a dataset that amalgamates CT images generated as \textit{fake} through CT-GAN with \textit{real} CT images sourced from the LIDC-IDRI dataset. Our training of 13 distinct DCNN models, guided by the minimization of binary cross-entropy loss, allowed us to uncover distinctive strengths inherent to these models.

Remarkably, ResNet50V2 emerged as the preeminent model, distinguished by exceptional precision and specificity. Its consistent and reliable performance profile signifies its efficacy in the context of medical image authentication, addressing the challenges posed by GAN-generated images. In contrast, DenseNet169 showcased notable excellence in accuracy, recall, and F1-score, highlighting its stability and effectiveness in navigating the nuanced distribution within CT scan images. Furthermore, our findings unveiled that MobileNetV3Large, while offering comparable performance to leading models, stands out due to its minimal parameters and swift inference time. For future work, we plan to further our experiments with a larger set of DCNNs and explore the effects of hyperparameter choice. We are also interested in investigating the causality behind the DCNN choice to select a certain label over another and have a strong explainability framework.

These insights collectively lay the groundwork for future research endeavors within the field of DCNN applications in medical imagery, particularly in the authentication of CT scans. Additionally, they provide a valuable resource for selecting an appropriate DCNN model tailored to the unique attributes of a given medical imaging scenario, aligning with the critical need for reliable approaches in the face of GAN-generated challenges.

{\small
\bibliography{egbib}

\begin{thebibliography}{10}

\bibitem{Garcea2023}
Fabio Garcea, Alessio Serra, Fabrizio Lamberti, and Lia Morra.
\newblock Data augmentation for medical imaging: A systematic literature review.
\newblock {\em Computers in Biology and Medicine}, 152:106391, 1 2023.

\bibitem{Osuala2021}
Richard Osuala, Kaisar Kushibar, Lidia Garrucho, Akis Linardos, Zuzanna Szafranowska, Stefan Klein, Ben Glocker, Oliver Diaz, and Karim Lekadir.
\newblock Data synthesis and adversarial networks: A review and meta-analysis in cancer imaging.
\newblock {\em Medical Image Analysis}, 84:102704, 2023.

\bibitem{2021paired}
Alaa Abu-Srhan, Israa Almallahi, Mohammad~AM Abushariah, Waleed Mahafza, and Omar~S Al-Kadi.
\newblock Paired-unpaired unsupervised attention guided gan with transfer learning for bidirectional brain mr-ct synthesis.
\newblock {\em Computers in Biology and Medicine}, 136:104763, 2021.

\bibitem{Mirsky2019}
Yisroel Mirsky, Tom Mahler, Ilan Shelef, and Yuval Elovici.
\newblock Ct-gan: Malicious tampering of 3d medical imagery using deep learning.
\newblock In {\em Proceedings of the 28th USENIX Conference on Security Symposium}, SEC'19, page 461–478, USA, 2019. USENIX Association.

\bibitem{Karras2018}
Tero Karras, Samuli Laine, and Timo Aila.
\newblock A style-based generator architecture for generative adversarial networks.
\newblock In {\em Proceedings of the IEEE/CVF conference on computer vision and pattern recognition}, pages 4401--4410, 2019.

\bibitem{Rana2021}
Md.~Shohel Rana, Beddhu Murali, and Andrew~H. Sung.
\newblock Deepfake detection using machine learning algorithms.
\newblock In {\em 2021 10th International Congress on Advanced Applied Informatics (IIAI-AAI)}, pages 458--463, 2021.

\bibitem{Badale2018}
Anuj Badale, Lionel Castelino, Chaitanya Darekar, and Joanne Gomes.
\newblock Deep fake detection using neural networks.
\newblock In {\em 15th IEEE international conference on advanced video and signal based surveillance (AVSS)}, 2018.

\bibitem{Khatri2023}
Nishika Khatri, Varun Borar, and Rakesh Garg.
\newblock A comparative study: Deepfake detection using deep-learning.
\newblock In {\em 2023 13th International Conference on Cloud Computing, Data Science \& Engineering (Confluence)}, pages 1--5, 2023.

\bibitem{simonyan2014very}
Karen Simonyan and Andrew Zisserman.
\newblock Very deep convolutional networks for large-scale image recognition.
\newblock {\em arXiv preprint arXiv:1409.1556}, 2014.

\bibitem{sandler2018mobilenetv2}
Mark Sandler, Andrew Howard, Menglong Zhu, Andrey Zhmoginov, and Liang-Chieh Chen.
\newblock Mobilenetv2: Inverted residuals and linear bottlenecks.
\newblock In {\em Proceedings of the IEEE conference on computer vision and pattern recognition}, pages 4510--4520, 2018.

\bibitem{chollet2017xception}
Fran{\c{c}}ois Chollet.
\newblock Xception: Deep learning with depthwise separable convolutions.
\newblock In {\em Proceedings of the IEEE conference on computer vision and pattern recognition}, pages 1251--1258, 2017.

\bibitem{szegedy2016rethinking}
Christian Szegedy, Vincent Vanhoucke, Sergey Ioffe, Jon Shlens, and Zbigniew Wojna.
\newblock Rethinking the inception architecture for computer vision.
\newblock In {\em Proceedings of the IEEE conference on computer vision and pattern recognition}, pages 2818--2826, 2016.

\bibitem{Rana2020}
Md.~Shohel Rana and Andrew~H. Sung.
\newblock Deepfakestack: A deep ensemble-based learning technique for deepfake detection.
\newblock In {\em 2020 7th IEEE International Conference on Cyber Security and Cloud Computing (CSCloud)/2020 6th IEEE International Conference on Edge Computing and Scalable Cloud (EdgeCom)}, pages 70--75, 2020.

\bibitem{Chen2021}
Hong-Shuo Chen, Mozhdeh Rouhsedaghat, Hamza Ghani, Shuowen Hu, Suya You, and C.-C. Jay~Kuo.
\newblock Defakehop: A light-weight high-performance deepfake detector.
\newblock In {\em 2021 IEEE International Conference on Multimedia and Expo (ICME)}, pages 1--6, 2021.

\bibitem{Ciftci2019}
Umur~Aybars Ciftci, Ilke Demir, and Lijun Yin.
\newblock Fakecatcher: Detection of synthetic portrait videos using biological signals.
\newblock {\em IEEE transactions on pattern analysis and machine intelligence}, 2020.

\bibitem{Chen2020}
Tianxiang Chen, Avrosh Kumar, Parav Nagarsheth, Ganesh Sivaraman, and Elie Khoury.
\newblock Generalization of audio deepfake detection.
\newblock In {\em Odyssey}, pages 132--137, 2020.

\bibitem{Guera2018}
David G{\"u}era and Edward~J Delp.
\newblock Deepfake video detection using recurrent neural networks.
\newblock In {\em 2018 15th IEEE international conference on advanced video and signal based surveillance (AVSS)}, pages 1--6. IEEE, 2018.

\bibitem{Zhou2021}
Yipin Zhou and Ser-Nam Lim.
\newblock Joint audio-visual deepfake detection.
\newblock In {\em Proceedings of the IEEE/CVF International Conference on Computer Vision}, pages 14800--14809, 2021.

\bibitem{Mittal2020}
Trisha Mittal, Uttaran Bhattacharya, Rohan Chandra, Aniket Bera, and Dinesh Manocha.
\newblock Emotions don't lie: An audio-visual deepfake detection method using affective cues.
\newblock In {\em Proceedings of the 28th ACM international conference on multimedia}, pages 2823--2832, 2020.

\bibitem{Vaswani2017}
Ashish Vaswani, Noam Shazeer, Niki Parmar, Jakob Uszkoreit, Llion Jones, Aidan~N Gomez, \L~ukasz Kaiser, and Illia Polosukhin.
\newblock Attention is all you need.
\newblock In I.~Guyon, U.~Von Luxburg, S.~Bengio, H.~Wallach, R.~Fergus, S.~Vishwanathan, and R.~Garnett, editors, {\em Advances in Neural Information Processing Systems}, volume~30. Curran Associates, Inc., 2017.

\bibitem{Dosovitskiy2020}
Alexey Dosovitskiy, Lucas Beyer, Alexander Kolesnikov, Dirk Weissenborn, Xiaohua Zhai, Thomas Unterthiner, Mostafa Dehghani, Matthias Minderer, Georg Heigold, Sylvain Gelly, et~al.
\newblock An image is worth 16x16 words: Transformers for image recognition at scale.
\newblock {\em arXiv preprint arXiv:2010.11929}, 2020.

\bibitem{Wang2022}
Junke Wang, Zuxuan Wu, Wenhao Ouyang, Xintong Han, Jingjing Chen, Yu-Gang Jiang, and Ser-Nam Li.
\newblock M2tr: Multi-modal multi-scale transformers for deepfake detection.
\newblock In {\em Proceedings of the 2022 International Conference on Multimedia Retrieval}, ICMR '22, page 615–623, New York, NY, USA, 2022. Association for Computing Machinery.

\bibitem{Ganguly2022}
Shreyan Ganguly, Aditya Ganguly, Sk~Mohiuddin, Samir Malakar, and Ram Sarkar.
\newblock Vixnet: Vision transformer with xception network for deepfakes based video and image forgery detection.
\newblock {\em Expert Systems with Applications}, 210:118423, 12 2022.

\bibitem{Budhiraja2022}
Rajat Budhiraja, Manish Kumar, M.K. Das, Anil~Singh Bafila, and Sanjeev Singh.
\newblock Medifaked: Medical deepfake detection using convolutional reservoir networks.
\newblock In {\em 2022 IEEE Global Conference on Computing, Power and Communication Technologies (GlobConPT)}, pages 1--6, 2022.

\bibitem{Solaiyappan2022}
Siddharth Solaiyappan and Yuxin Wen.
\newblock Machine learning based medical image deepfake detection: A comparative study.
\newblock {\em Machine Learning with Applications}, 8:100298, 6 2022.

\bibitem{Armato2011}
Armato iii, s. g., mclennan, g., bidaut, l., mcnitt-gray, m. f., meyer, c. r., reeves, a. p., zhao, b., aberle, d. r., henschke, c. i., hoffman, e. a., kazerooni, e. a., macmahon, h., van beek, e. j. r., yankelevitz, d., biancardi, a. m., bland, p. h., brown, m. s., engelmann, r. m., laderach, g. e., max, d., pais, r. c. , qing, d. p. y. , roberts, r. y., smith, a. r., starkey, a., batra, p., caligiuri, p., farooqi, a., gladish, g. w., jude, c. m., munden, r. f., petkovska, i., quint, l. e., schwartz, l. h., sundaram, b., dodd, l. e., fenimore, c., gur, d., petrick, n., freymann, j., kirby, j., hughes, b., casteele, a. v., gupte, s., sallam, m., heath, m. d., kuhn, m. h., dharaiya, e., burns, r., fryd, d. s., salganicoff, m., anand, v., shreter, u., vastagh, s., croft, b. y., clarke, l. p. (2015). \textbf{Data From LIDC-IDRI} [data set]. the cancer imaging archive.

\bibitem{Isola2017}
Phillip Isola, Jun-Yan Zhu, Tinghui Zhou, and Alexei~A. Efros.
\newblock Image-to-image translation with conditional adversarial networks.
\newblock In {\em 2017 IEEE Conference on Computer Vision and Pattern Recognition (CVPR)}, pages 5967--5976, 2017.

\bibitem{liu2022convnet}
Zhuang Liu, Hanzi Mao, Chao-Yuan Wu, Christoph Feichtenhofer, Trevor Darrell, and Saining Xie.
\newblock A convnet for the 2020s.
\newblock In {\em Proceedings of the IEEE/CVF Conference on Computer Vision and Pattern Recognition}, pages 11976--11986, 2022.

\bibitem{huang2017densely}
Gao Huang, Zhuang Liu, Laurens Van Der~Maaten, and Kilian~Q Weinberger.
\newblock Densely connected convolutional networks.
\newblock In {\em Proceedings of the IEEE conference on computer vision and pattern recognition}, pages 4700--4708, 2017.

\bibitem{tan2019efficientnet}
Mingxing Tan and Quoc Le.
\newblock Efficientnet: Rethinking model scaling for convolutional neural networks.
\newblock In {\em International conference on machine learning}, pages 6105--6114. PMLR, 2019.

\bibitem{tan2021efficientnetv2}
Mingxing Tan and Quoc Le.
\newblock Efficientnetv2: Smaller models and faster training.
\newblock In {\em International conference on machine learning}, pages 10096--10106. PMLR, 2021.

\bibitem{howard2019searching}
Andrew Howard, Mark Sandler, Grace Chu, Liang-Chieh Chen, Bo~Chen, Mingxing Tan, Weijun Wang, Yukun Zhu, Ruoming Pang, Vijay Vasudevan, et~al.
\newblock Searching for mobilenetv3.
\newblock In {\em Proceedings of the IEEE/CVF international conference on computer vision}, pages 1314--1324, 2019.

\bibitem{radosavovic2020designing}
Ilija Radosavovic, Raj~Prateek Kosaraju, Ross Girshick, Kaiming He, and Piotr Doll{\'a}r.
\newblock Designing network design spaces.
\newblock In {\em Proceedings of the IEEE/CVF conference on computer vision and pattern recognition}, pages 10428--10436, 2020.

\bibitem{he2016identity}
Kaiming He, Xiangyu Zhang, Shaoqing Ren, and Jian Sun.
\newblock Identity mappings in deep residual networks.
\newblock In {\em Computer Vision--ECCV 2016: 14th European Conference, Amsterdam, The Netherlands, October 11--14, 2016, Proceedings, Part IV 14}, pages 630--645. Springer, 2016.

\bibitem{tensorflow2015-whitepaper}
Mart\'{i}n Abadi, Ashish Agarwal, Paul Barham, Eugene Brevdo, Zhifeng Chen, Craig Citro, Greg~S. Corrado, Andy Davis, Jeffrey Dean, Matthieu Devin, Sanjay Ghemawat, Ian Goodfellow, Andrew Harp, Geoffrey Irving, Michael Isard, Yangqing Jia, Rafal Jozefowicz, Lukasz Kaiser, Manjunath Kudlur, Josh Levenberg, Dandelion Man\'{e}, Rajat Monga, Sherry Moore, Derek Murray, Chris Olah, Mike Schuster, Jonathon Shlens, Benoit Steiner, Ilya Sutskever, Kunal Talwar, Paul Tucker, Vincent Vanhoucke, Vijay Vasudevan, Fernanda Vi\'{e}gas, Oriol Vinyals, Pete Warden, Martin Wattenberg, Martin Wicke, Yuan Yu, and Xiaoqiang Zheng.
\newblock {TensorFlow}: Large-scale machine learning on heterogeneous systems, 2015.
\newblock Software available from tensorflow.org.

\bibitem{Deng2009imagenet}
Jia Deng, Wei Dong, Richard Socher, Li-Jia Li, Kai Li, and Li~Fei-Fei.
\newblock Imagenet: A large-scale hierarchical image database.
\newblock In {\em 2009 IEEE conference on computer vision and pattern recognition}, pages 248--255. Ieee, 2009.

\bibitem{kingma2014adam}
Diederik~P Kingma and Jimmy Ba.
\newblock Adam: A method for stochastic optimization.
\newblock {\em arXiv preprint arXiv:1412.6980}, 2014.

\bibitem{Rajpurkar2017}
Pranav Rajpurkar, Jeremy Irvin, Kaylie Zhu, Brandon Yang, Hershel Mehta, Tony Duan, Daisy Ding, Aarti Bagul, Curtis Langlotz, Katie Shpanskaya, et~al.
\newblock Chexnet: Radiologist-level pneumonia detection on chest x-rays with deep learning.
\newblock {\em arXiv preprint arXiv:1711.05225}, 2017.

\bibitem{yang2018netadapt}
Tien-Ju Yang, Andrew Howard, Bo~Chen, Xiao Zhang, Alec Go, Mark Sandler, Vivienne Sze, and Hartwig Adam.
\newblock Netadapt: Platform-aware neural network adaptation for mobile applications.
\newblock In {\em Proceedings of the European Conference on Computer Vision (ECCV)}, pages 285--300, 2018.

\end{thebibliography}
}
\section*{Appendix}
\appendix

\renewcommand{\thefigure}{A\arabic{figure}}
\renewcommand{\thetable}{A\arabic{table}}
\setcounter{figure}{0}   
\setcounter{table}{0}  

\section{Model Selection}\label{selection}

Based on our analysis, as highlighted in the CheXNet model presented by \cite{Rajpurkar2017}, the performance of architectures that have demonstrated effectiveness on ImageNet \cite{Deng2009imagenet} does not necessarily translate to superior results with CheXNet. CheXNet, specifically trained on chest X-ray images, challenges the assumption that models excelling on general datasets, such as ImageNet, will perform equally well on domain-specific data like medical images. Therefore, we derived that existing DCNNs that perform greatly with generic data might not also perform as well in our case. As a result, we have decided to explore alternative well-established DCNN architectures to address this challenge.

To this end, we determined that the models we chose to experiment with are most reasonable to satisfy two conditions: a) experiment with various model families by choosing the model with the largest size from each family, but that is restricted by $2$. The model we choose should not exceed $30$M parameters due to computational constraints. Given that the CheXNet paper claimed intra-family generalization, we decided to experiment with all of the models in the DenseNet family to test the claim.

To choose which models we will perform 3 runs to validate their stability, we will need to calculate the harmonic mean of accuracy, F1-Score, and specificity as mentioned before. The formula for the harmonic mean H given N numbers is:
\begin{equation}
\text{H} =  \frac{n}{\sum_{i=1}^{n}\frac{1}{x_i}} \label{H}
\end{equation}
For the case of our experiment, the formula is:
\begin{equation}
\text{H} =  \frac{3}{\frac{1}{\text{Accuracy}} + \frac{1}{\text{F1-Score}} + \frac{1}{\text{Specificity}}}\label{H2}
\end{equation}
Table \ref{htable} shows the result of H for all models. Given the results of H and after reducing the number of models, we will proceed with experimenting with two more runs with the winner models, i.e. DenseNet121, DenseNet169, EfficientNetB4, EfficientNetV2S, ResNet50V2, and Xception, then calculate the mean and standard deviations of the result, consequently analyzing their results.
\begin{table}
\caption{Results of the harmonic mean of the accuracy, F1-Score, and specificity on the evaluation metrics results of the proposed models after one run. The best 6 models are in bold.}\label{htable}
\begin{tabular*}{\columnwidth}{@{\extracolsep\fill}lcc}
\toprule%
Model & H \\
\midrule
ConvNeXtTiny & 0.9519 \\
DenseNet121 & \textbf{0.9595} \\
DenseNet169 & \textbf{0.9752} \\
DenseNet201 & 0.9538 \\
EfficientNetB4 & \textbf{0.9678} \\
EfficientNetV2S & \textbf{0.9661} \\
InceptionV3 & 0.9468 \\
MobileNetV3Large & 0.9557 \\
RegNetX040 & 0.9529 \\
RegNetY040 & 0.9486 \\
ResNet50V2 & \textbf{0.9780} \\
VGG19 & 0.9517 \\
Xception & \textbf{0.9659} \\
\bottomrule
\end{tabular*}
\end{table}

\renewcommand{\thefigure}{B\arabic{figure}}
\renewcommand{\thetable}{B\arabic{table}}
\setcounter{figure}{0}   
\setcounter{table}{0}  
\section{Evaluation Metrics}\label{eval}
To calculate the evaluation metrics, we describe the four possible outcomes of our models, namely, True Positives (TPs), True Negatives (TNs), False Positives (FPs), and False Negatives (FNs).
\\We get a TP when the model correctly predicts the positive class, a TN when the model correctly predicts the negative class, an FP when the model incorrectly predicts the positive class, and finally, an FN when the model incorrectly predicts the negative class.

\textbf{Accuracy.} Accuracy measures the proportion of correct predictions made by the model out of all the predictions made. In other words, accuracy tells us how often the model's predictions were correct. The formula for accuracy is:

\begin{equation}
\text{Accuracy} = \frac{TP + TN}{TP + TN + FP + FN}\label{accuracy}
\end{equation}
\\\textbf{Precision.} Precision measures the proportion of true positive predictions out of all the positive predictions made by the model. In other words, precision tells us how many of the positive predictions made by the model were correct. The formula for precision is:

\begin{equation}
\text{Precision} = \frac{TP}{TP + FP}\label{precision}
\end{equation}
\\\textbf{Recall.} Recall measures the proportion of true positive predictions out of all the actual positive instances in the dataset. In other words, recall tells us how many of the positive instances in the dataset were correctly identified by the model. Recall is also called sensitivity or true positive rate (TPR). The formula for recall is:

\begin{equation}
\text{Recall} = \frac{TP}{TP + FN}\label{recall}
\end{equation}
\\\textbf{Specificity.} Specificity measures the proportion of true negative predictions out of all the actual negative instances in the dataset. In other words, specificity tells us how many of the negative instances in the dataset were correctly identified by the model as negative. Specificity is also known as selectivity or true negative rate (TNR). The formula for specificity is:

\begin{equation}
\text{Specificity} = \frac{TN}{TN + FP}\label{specificity}
\end{equation}
\\\textbf{F1-score.} F1-score is the harmonic mean of precision and recall, which provides a single measure of a model's accuracy by balancing the trade-off between precision and recall. The formula for F1-score is:

\begin{equation}
\text{F1-Score} = \frac{2 \times \text{Precision} \times \text{Recall}}{\text{Precision} + \text{Recall}}\label{f1}
\end{equation}
\\\textbf{Fall-out.} Fall-out measures the proportion of actual negative instances that were incorrectly classified as positive by the model. Fall-out is also known as the false positive rate. The formula for fall-out is:

\begin{equation}
\text{Fall-out} =  \frac{FP}{FP + TN}\label{fpr}
\end{equation}
We will not directly mention the fall-out in the results table, but we will rather use it to calculate the area under the curve (AUC).
\\\textbf{AUC.} AUC stands for the Area Under the Curve, specifically the ROC curve. The ROC curve (Receiver Operating Characteristic curve) is a plot of the true positive rate (sensitivity) against the false positive rate (fall-out) for various classification thresholds. AUC represents the area under the ROC curve, which ranges from 0 to 1, with higher values indicating better performance of the model. The formula for AUC is:

\begin{equation}
\text{AUC} =  \int_{0}^{1} \text{ROC}(x) \, dx\label{auc}
\end{equation}

\end{document}